\documentclass[11pt]{article}
\pdfoutput=1

\usepackage{arxiv}
\usepackage[utf8]{inputenc}
\title{Transfer training from smaller language model}
\author{Han Zhang}
\date{April 2021}
\usepackage{natbib}
\usepackage{graphicx}
\begin{document}
\maketitle
\section{ABSTRACT}
Large language models have led to state-of-the-art accuracies across a range of tasks. However,training large language model needs massive computing resource, as more and more open source pre-training models are available, it is worthy to study how to take full advantage of available model. We find a method to save training time and resource cost by changing the small well-trained model to large model. We initialize a larger target model from a smaller source model by copy weight values from source model and padding with zeros or small initialization values on it to make the source and target model have approximate outputs, which is valid due to block matrix multiplication and residual connection in transformer structure. We test the target model on several data sets and find it is still comparable with the source model. When we continue training the target model, the training loss can start from a smaller value.

\section{INTRODUCTION}

Transformer~\cite{transformer} is a simple network architecture proposed by Vaswani to solving machine translation problem, it is based solely on attention mechanisms and soon become one of the most popular model structure in deep learning community. The early applications of transformer is Natural Language Processing(NLP) such as machine translation~\cite{Jiacheng-translation}. After Open-AI and Google successively propose large transformer-based pretrained Language Models(LM) -- GPT~\cite{GPT1} and BERT ~\cite{bert}, scholars study on all kinds of NLP tasks by using pretrained LM, including Text Classification, Named Entity Recognition, Natural Language Inference, Read Comprehension, Automated Abstracting and so on. Except NLP tasks, transformers also be applied for many multi-model tasks to bridging text and video\cite{BridgingTextandVideo}, time-series forecasting tasks~\cite{Informer} and so on. As the success of the pre-training model, there are two routes to enhance model's performance. The first is Lightweighting the pretrained model such as model compressing~\cite{EdgeBERT,albert} or accelerating the inference~\cite{Inference}, the second is increasing model size, such as GPT3~\cite{gpt3} and Switch Transformer ~\cite{switchTransformer}. 

Megatron~\cite{Megatron} is the first open-sourced project to increasing transformer size by efficient intra-layer model parallel approach that enables training transformer models with 8.3 billion of parameters on 512 GPUs. T5~\cite{T5} increasing the parameters number to 11 billion by Text-to-Text Transfer Transformer, and treat every text processing problem as a “text-to-text” problem, witch shows some general intelligence ability. The study of GPT3~\cite{gpt3} shows that large Language Model has strong context learning ability, which without parameters updating but are compareble with state of the art fine-tune style methods. And as the model size increasing, there is still room for performerence improving. Lots of work shows that larger language models showing potential to more intelligence ablity~\cite{cpm}, however, training a large model form head costs much resources for instance that training GPT3 from the beginning costs more than ten millions of dollars.

There have been some studies on making pretraining more efficiency by improving pretraining tasks. Roberta ~\cite{RoBERTa} employs a dynamic mask strategy to make the best of each sentence in corpus witch reduces the demand for corpus volume. Electra ~\cite{Electra} uses several efficient pre-training objectives for transformers based models by adversarial learning, which requires smaller classification heads and exhibits a general superiority over MLM. PMLM ~\cite{PMLM} adopts a probabilistic masking scheme for the MLM, and has a unique ability to autoregressively generate sequences in arbitrary order, which improves training effctively by enhance the difficulty of the pretraining task. 

Efficient Transformers is a research focus to reducing transformer complexity to saving training resource cost. The typical approachs are Memory method, Low-Rank method and Kernel method. Memory method is to leverage a side memory module that can access multiple tokens at once, for instance Longformer~\cite{Longformer} and set transformer~\cite{SetTransformer}. Low-Rank method is to improve efficiency by leveraging low-rank approximations of the self-attention matrix such as Linformer~\cite{Linformer}. Kernel method to improve the efficiency of Transformers is to view the attention mechanism through kernelization such as  Performer~\cite{Performer}, Linformer~\cite{Linformer} and Linear Transformer~\cite{LinearTransformer}. Yi Tay~\cite{LongRangeArena} proposes a long range arena to evaluate the Efficient Transformers' performance, speed and memory footprint.

Vertical to improving pre-training tasks or optimizing transformer model, we study on transfering model parameters and it can be compatible with all of above methods. Our works focus on how to reduce the resource cost of training large language model from the beginning. There are many open-resourced pretrained models in transformers library~\cite{wolf-etal-2020-transformers} such as BERT, BART, XLM, ALBERT, Electra, XLNet and so on. In the common situation, we fetch the pretrained model and add task-specific layers on it then fine-tune the whole model in different tasks. Continuing this idea, we could also add new parameters to each layer to broaden the model or add new layers to deepen the model. According to the rule of block matrix multiplication , when we padding zeros at matrix in the same position, the result of matrix multiplication remain unchanged. The details will be discussed in section 4.

\section{RELATED WORK}
In the following paragraphs we summarize the related work needed to introduce our approach.
Transformer is a sequence-to-sequence model and therefore comprises an encoder and a decoder. 

\textbf{Encoder} The Transformer’s encoder is a function defined as the composition of L identical layers or blocks, each composed of two sub-layers. The first sub-layer is the self-attention
mechanism which allows the encoder to focus on the relevant part of the sequence for each position, similarly to the inter attention depicted in Figure \ref{Tab:Attention-ori}. The second sub-layer is a simple fully connected feed-forward network applied independently and identically to every position (position-wise). The feedforward network increases the encoder’s expressiveness and transforms the self-attention’s output for the next layer.

\textbf{Stacked Encoder}: BERT is a stacked encoder Transformer, which inputs a sequence of tokens and applies position and token and token type embeddings. Each layer applies multi-head self-attention in combination with a feedforward network, layer normalization, and residual connections. The BERT base model has 12 layers and 12 heads.

\textbf{Decoder} The decoder is also composed of L identical layers. Note that although it is common for the decoder to have the same number of layers as the encoder, one may adjust their depth independently. Each decoder’s layer comprises three sub-layers. The first sub-layer is the self-attention mechanism as in the encoder, except that future positions are masked. Indeed, the encoder is allowed to look at future positions since the input sequence is entirely available. The decoder, however, cannot look at future positions since they have not yet been predicted. Therefore, the i-th position may only attend to positions less than i. The second sub-layer is the inter-attention mechanism, which helps the decoder focus on the relevant parts of the input, such as depicted in Figure 3. Finally, the third sub-layer is a simple feed-forward network. As for the encoder, a residual connection and a layer normalization are applied to each sub-layer.

\textbf{Stacked Decoder}: GPT-2 is a stacked decoder Transformer, which inputs a sequence of tokens and applies position and token embeddings followed by several decoder layers. Each layer applies multi-head attention combination with a feedforward network, layer normalization, and residual connections. And the attention score is controlled by a lower triangular matrix which is designed to prevent the current word from seeing the following words. The GPT-2 small model has 12 layers and 12 heads.

To verify the universality of the method, we apply our method to transfer BERT and GPT models, both of which are typical transformer based pretrained language model.

\section{Method: transformer-transfering}

In this section, we present the implementation of Weight-transferring. Given a well-trained transformer model such as GPT-small as our source model, there are two kinds of variables in it --- parameters and buffers. Parameters are variables can be updated by optimizer, including embedding weight, dense weight and bias, buffers are variables can’t be updated by optimizer, such as bias values in attention layer which is used to control attention score in masked position.

\begin{figure}[!ht]
\centering
\includegraphics[scale=0.4]{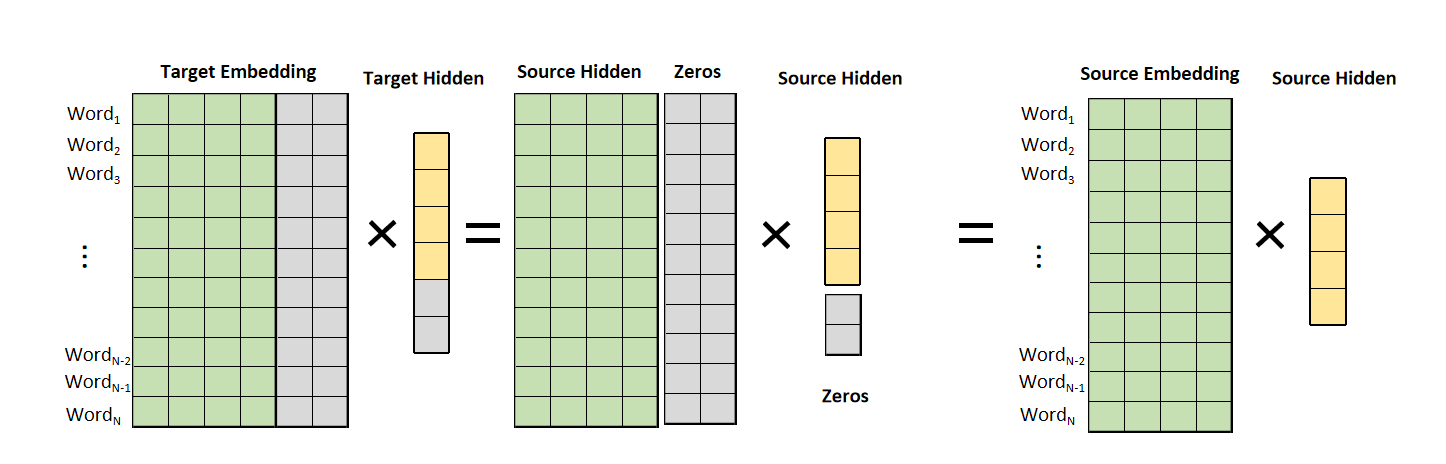}
\caption{Embedding transfer}
\label{Tab:Embdeeing}
\end{figure}

\textbf{Embedding-transfer.} Different to directly copying word embedding matrix like word2vector, we need to transfer a smaller size matrix from source model to a bigger matrix in target model. For computational equivalence, we split the target embedding matrix into two blocks by rows, the first block has the same size with source embedding matrix, so we can straightforward copy the weights from source. The second block is initialized by zeros tensor (showed in Figure \ref{Tab:Embdeeing}) so that the inner product operation results in the language model head (the last layer for GPT) can be unchanged.

\begin{figure}[!ht]
\centering
\includegraphics[scale=0.4]{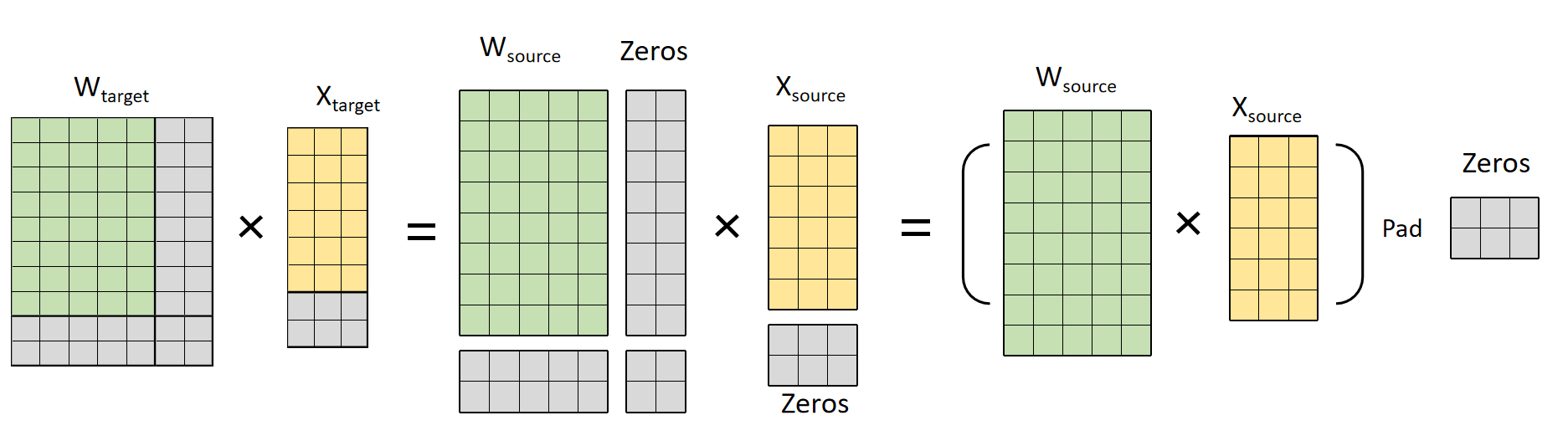}
\caption{ Dense-transfer}
\label{Tab:Dense}
\end{figure}

\textbf{Dense-transfer.} In order of calculation in transformer, there are several mapping layers after embedding and LayerNorm layer. Such as in self-attention layer, the hidden states matrix are map into three time width matrix for query, key, and value to calculating attention score. If the dimension of hidden states of target model is 1024, then there is a (3072,1024) shape weight and a (3027,) shape bias parameters in the mapping layer. If the dimension of source model is 768, so the matrix shape in mapping layer is (2304,768) which both dimension are not same as matrix in target.  Block Matrix Multiplication ensure our transfer method is valid. As showed in  Figure \ref{Tab:Dense}, by padding zeros at the tail of source matrix, the result of Block Matrix Multiplication is equal to directly do matrix multiplication on source matrix and then padding zeros at the tail. And the bias transfer is analogical. 

\begin{figure}[!ht]
\centering
\includegraphics[scale=0.4]{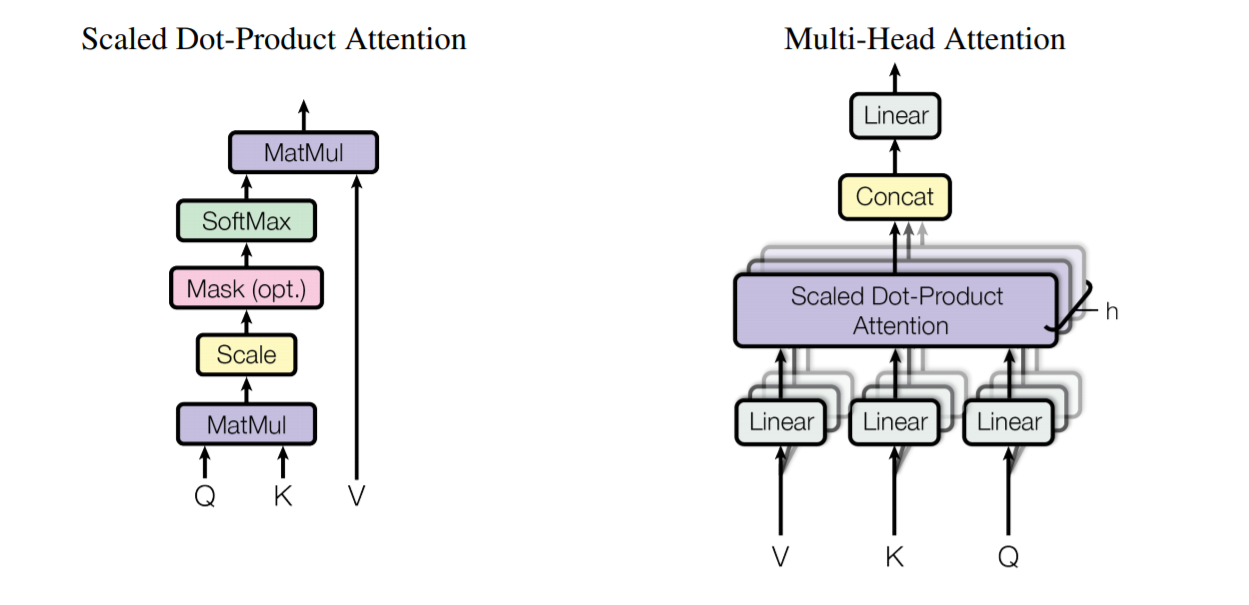}
\caption{Attention structure}
\label{Tab:Attention-ori}
\end{figure}

\textbf{Attention Layer transfer.}  As showed in Figure \ref{Tab:Attention-ori}, attention layer calculation is consist of several Matrix Multiplication, so we can use the same method as Dense transfer. The multi-head attention splits the hidden states into multi-heads and execute scaled dot-product attention respectively. But if we change the dimension of hidden states or head number, each head will change and may influence the attention score. There are two ways to ensure the results valid. The first way is keep each head dimension and only increase the head numbers, the second way is padding each head zeros at the tail respectively, both of them are mathematically equivalent.

\begin{figure}[!ht]
\centering
\includegraphics[scale=0.55]{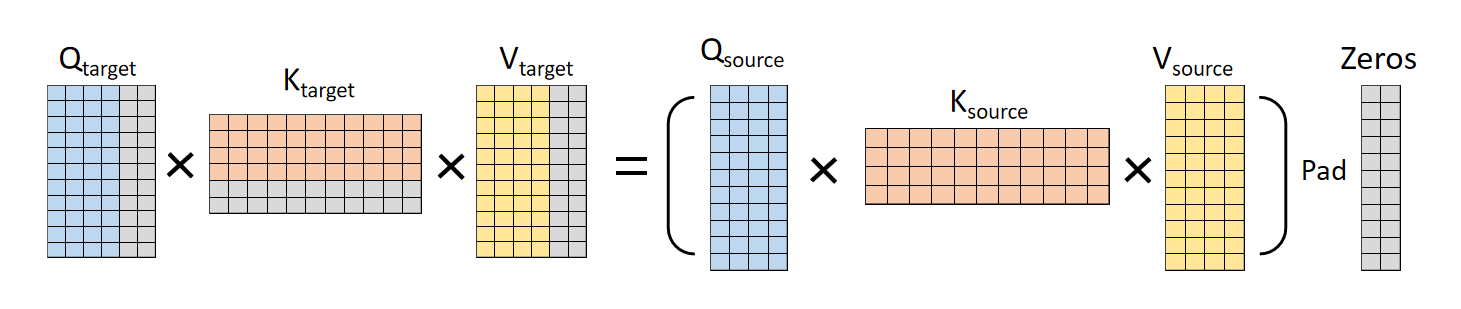}
\caption{ Attention-transfer}
\label{Tab:Attention}
\end{figure}

Figure \ref{Tab:Attention} shows the scaled dot-product attention calculation process of one head. 

\textbf{About LayerNorm.}  Layer normalization (LayerNorm) is a technique to normalize the distributions of intermediate layers. It enables smoother gradients, faster training, and better generalization accuracy, because of its capability in handling re-centering and re-scaling of both inputs and weight matrix. However, both re-centering and re-scaling opeartion are related to hidden size, which is changed after transfering.  It is not totally mathematics equivalent for source and target model, but the influence is very small  since the scaling weight and bias can be updated fast after training several steps to adjust new model parameters.  

\textbf{Deeper Layers Parameters.}  Deepening neural net is the the most commonly way to increasing model size. When we transfer a source model with few layers to a deeper target model, the parameters in deep layers need to be initialized in smaller values. Because of residual connection, the inputs and outputs of these layers are approximate. It makes the target model have similar output distribution with the source model.

\section{Experiments}
In this section, we conduct extensive experiments on inference performance and subsequent training. We test BERT and GPT on Cloze, Next Sentence Prediction and Next Word Prediction task, which are pretraining objective in the stage of pre-training. Then we continue train the target model find it can reach better performance than source model.

\subsection{Infertence ability}
In this subsection, we test our target BERT model on LAMA~\cite{petroni2020how} dataset, which is a probe for analyzing the factual and commonsense knowledge contained in pretrained language models. As can be seen in Table \ref{tab:Mask}, after transferring, the target model still remain the mask filling ability, which shows that our transfer method is valid. The performance loss is due to the LayerNorm part, which is not mathematically equivalent when transferring.

\begin{table}
\centering
\renewcommand\arraystretch{1.25}
\caption{Mask Filling Task}
\label{tab:Mask}
\begin{tabular}{c|c|c} 

\hline
Dataset        & Source Model(108M) & Target Model(355M) \\
\hline
ConceptNet     & 14.80 & 7.20 \\
\hline
Squad          & 15.89 & 10.26 \\

\hline

\end{tabular}
\end{table}

\subsection{Subsequent training}
\begin{figure}[!ht]
\centering
\includegraphics[scale=0.52]{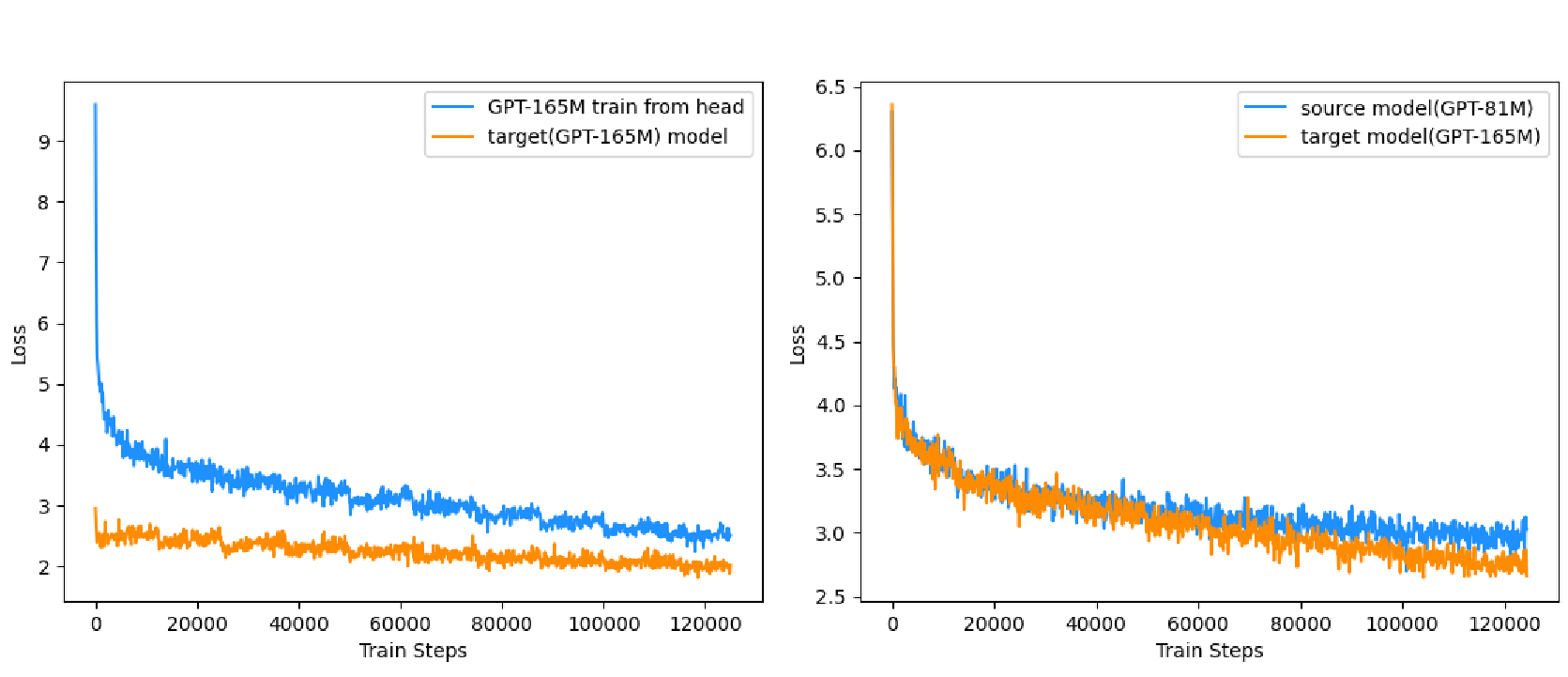}
\caption{Left is training loss compared on chat corpus, right is on douban corpus }
\label{Fig:two_plot}
\end{figure}

In this part, we train a 81 million parameters source dialog GPT~\cite{zhang2019dialogpt}~\cite{zhang2019dialogpt} model in chat corpus, then we transfer the source GPT model to a 165 million target model.  
Next, we conduct two experiments to verify our method. In the first experiment, We train the transfered target model and a random initialized 165 million GPT model on chat corpus. The training loss is showed in Figure \ref{Fig:two_plot} left, the target model have smaller loss at beginning steps.
In second experiment, we continue train the source and target model on douban corpus which is new to both models. As showed in Figure \ref{Fig:two_plot} right, the target model can achieve lower training loss.


\section{Conclusion}
We propose a transfer strategy which can increase the model size with acceptable performance decreasing by transferring the parameters of source model. It is compute resource saving by avoid training a large model from the beginning. The method is valid due to block matrix multiplication and residual connection in transformer structure except the LayerNrom layer which causes a little performance decreasing. Our feature work is optimizing the transfer strategy to more compatibility and with less performance decreasing.

\bibliographystyle{unsrtnat}
\bibliography{paper}

\section{Appendix}
\begin{figure}[!ht]
\centering
\includegraphics[scale=0.52]{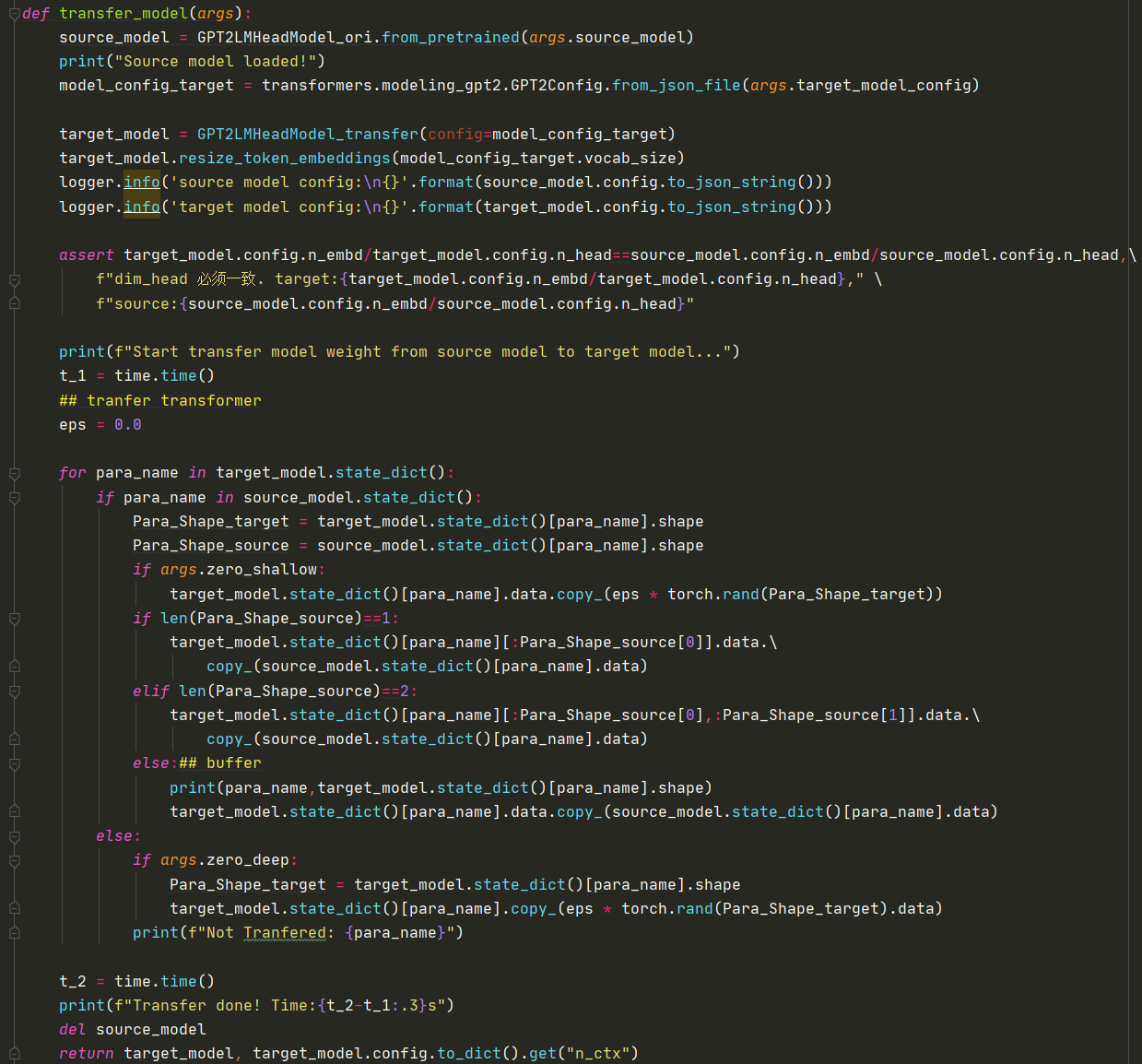}
\caption{Code for transferring GPT}
\label{Tab:code}
\end{figure}

The core codes of transferring gpt model is show in Figure \ref{Tab:code}

\end{document}